\def\year{2022}\relax
\documentclass[letterpaper]{article} %
\usepackage{aaai22}  %
\usepackage{times}  %
\usepackage{helvet}  %
\usepackage{courier}  %
\usepackage[hyphens]{url}  %
\usepackage{graphicx} %
\usepackage{natbib}  %
\usepackage{caption} %
\DeclareCaptionStyle{ruled}{labelfont=normalfont,labelsep=colon,strut=off} %
\usepackage{algorithm}

\usepackage{booktabs}
\usepackage{multirow}

\usepackage{newfloat}
\usepackage{subfig}
\usepackage{listings}
\floatstyle{ruled}
\newfloat{listing}{tb}{lst}{}
\floatname{listing}{Listing}

\title{\title{DocBed: A Multi-Stage OCR Solution for Documents with Complex Layouts}}
\author {
    Wenzhen Zhu,
    Negin Sokhandan,
    Guang Yang,
    Sujitha Martin,
    Suchitra Sathyanarayana
}
\affiliations {
    Amazon Web Services AI\\
    {
    \{wenzhu, ngnsl, yaguan, sujimart, suchisat\}@amazon.com
    }
}

\usepackage{bibentry}

\begin{document}

\maketitle

\begin{abstract}
Digitization of newspapers is of interest for many reasons including preservation of history, accessibility and search ability, etc. While digitization of documents such as scientific articles and magazines is prevalent in literature, one of the main challenges for digitization of newspaper lies in its complex layout (e.g. articles spanning multiple columns, text interrupted by images) analysis, which is necessary to preserve human read-order. This work provides a major breakthrough in the digitization of newspapers on three fronts: first, releasing a dataset of 3000 fully-annotated, real-world newspaper images from 21 different U.S. states representing an extensive variety of complex layouts for document layout analysis; second, proposing layout segmentation as a precursor to existing optical character recognition (OCR) engines, where multiple state-of-the-art image segmentation models and several post-processing methods are explored for document layout segmentation; third, providing a thorough and structured evaluation protocol for isolated layout segmentation and end-to-end OCR.

\end{abstract}

\section{Introduction}
In the digital age, information that used to be spread by printouts is disseminated at unforeseen speeds through digital formats. In parallel to the inventions of new types of media, an increasing number of archives and libraries are trying to create digital repositories with new technologies \cite{DBLP:journals/firstmonday/Cox07}. Digitization allows for preservation by creating an accessible surrogate, while at the same time enabling easier storage, indexing, and faster search.

The objective of this work is to build and investigate systems that can generalize well to the digitization of historical newspapers, where digitization here refers to extracting content (e.g. text, image, ads) in human read-order.
This is challenging because there are many variations among old newspapers, including font sizes and styles, number of columns, column width, column separators, advertisement placement, image placements, table style, break in continuity of articles across multiple columns, text versus non-text ratio, etc., to name a few. To address these challenges, this paper makes three contributions:
\begin{itemize}
    \item a new dataset of 3000 scanned pages sampled from 21 different newspaper publications spread across the U.S. states over a few decades and annotated with bounding boxes for seven categories.
    \item an end-to-end framework for document layout analysis that uses a variety of advanced segmentation and detection techniques, such as Mask R-CNN from Detectron2 \cite{wu2019detectron2} and Segmentation Transformer \cite{Zheng_2021_CVPR}, and several post-processing algorithms (that optimize the number of API calls and improve text recognition metrics) to handle various context-aware text extraction tasks, as a precursor to existing OCR engines like Amazon Textract and Tesseract.
    \item two benchmarking pipelines that allow evaluation of isolated layout segmentation and end-to-end text recognition.
\end{itemize}

\section{Related Work}
Document layout analysis is a critical step in document digitization and is often split into two tasks: page segmentation and region classification. The objective is to infer the underlying structure of an unstructured digital document and parse it into structured formats so that each article can be properly indexed. Previous work on this topic can be mainly grouped into two categories: (1) layout analysis methods and (2) data collection and layout label generation aiming to include larger varieties of document types. 

Various rule-based algorithms have been proposed and can be divided into bottom-up and top-down approaches. Bottom-up approaches first classify small components of an image and then cluster similar components to form regions. Top-down approaches cut the image recursively in vertical and horizontal directions along white spaces or boundary lines. \cite{DBLP:conf/icdar/Smith09} used a bottom-up method first to generate initial data-type hypothesis and then applied a top-down method to impose structural regions, which later became the layout analysis component of the Tesseract engine. \cite{DBLP:journals/jodl/KlampflGJK14} applied a bottom-up approach to first extract text blocks and then form geometrical relations via graph-based approaches between these blocks.

However, as these methods depend highly on heuristics and strategic thresholds, they are often greedy and sub-optimal. Some globally optimized methods were proposed to increase the flexibility of the segmentation methods to adapt the local thresholds to local variations. \cite{DBLP:conf/icdar/AgrawalD09a}
proposed a global page segmentation method based on Voronoi and Docstrum features.
\cite{DBLP:conf/icvgip/DasigiJJ08} formulated the document segmentation into an optimal spectral partitioning problem, for which a closed-form solution can be achieved instead of classic adhoc solutions. Lately, the applicability of deep neural networks methods has gained a lot of attention. Fully convolutional networks (FCN) has been explored in multiple studies \cite{DBLP:conf/icfhr/OliveiraSK18} \cite{DBLP:conf/das/WickP18} \cite{DBLP:conf/icdar/MeierSSAC17}. Later, proposals were made to include textual features to achieve better and finer page segmentation and region classification \cite{DBLP:journals/corr/abs-2002-06144} \cite{DBLP:conf/emnlp/KattiRGBBHF18}. 

A series of efforts have been made in the ICDAR challenges, led by researchers in the Pattern Recognition and Image Analysis (PRImA) Research Lab, to increase the diversity of document types for layout analysis. \cite{DBLP:conf/icdar/AntonacopoulosBPP09} created a dataset with a wide selection of complex and contemporary documents. The dataset consisted of 1240 images with corresponding accurate ground-truth and extensive metadata. However, particular emphasis was placed on magazines and technical journals. Old documents pose unique challenges compared to the contemporary documents due to lower print qualities and more complex layouts. To accommodate the need for digitizing historical document archives, \cite{DBLP:conf/icdar/ClausnerPPA15} presented a dataset comprising over 500 images of newspapers mostly published around the World War I period.
To get around the dataset size limit due to manual annotations requirements, \cite{DBLP:conf/icdar/ZhongTJ19} developed PubLayNet dataset by automatically matching XML representations with over 1 million article pdfs publicly available on PubMed Central. However, this method only applies to contemporary documents with XML representations.

\section{Datasets}
Two data sources are used in this work. The first is proposed as a part of our contributions and is primarily used for page layout segmentation analysis and benchmarking. The second is an existing dataset used for benchmarking the benefits of page layout segmentation on OCR results.

\paragraph{NewsNet7: Newspaper Layout Segmentation Dataset}
For layout segmentation analysis and benchmarking, 3000 images are sampled from an online database hosted at Chronicling America, produced by the National Digital Newspaper Program (NDNP), and annotated for page layout segments to create a new dataset called NewsNet7. Chronicling America is a website containing an extensive database of historic newspapers spanning all 50 states of the U.S. from the late 1700s to the mid-1900s. In order to capture a wide variety of complex layouts, NewsNet7 is sampled from different newspaper publishers spread across 21 different U.S. states. We also ensured each sample met a reasonable resolution (i.e. at least 4 megabytes) to maintain a certain degree of consistency across the samples. In addition, we also assure the sampled newspapers were covered uniformly within the period of their respective publisher and represented different pages of an issue. 
The 3000 selected newspaper images were manually annotated for page layout segmentation, in the form of bounding boxes for each of the following categories:
\begin{itemize}
    \item \textbf{Header}: Newspaper title related headers, usually appears at the very top.
    \item \textbf{Article title}: The title of each individual article which is followed by article body. 
    \item \textbf{Article body}: A body of text which follows an article title. Article body may span multiple columns and if so, annotations are broken into multiple bounding boxes, where each bounding box is restricted to span only one column.
    \item \textbf{Advertisement}: Text sections with visually more artistic and irregular formatting (e.g. different font style and size). To discern between ad and article, only visual cues were used and no consideration was given to content.
    \item \textbf{Image}: Represents art, photo, illustrations, etc.
    \item \textbf{Table}: Arrangement of data in rows and columns (normally comes in key-value pairs). 
    \item \textbf{Other}: Is assigned to segments not falling into any of the above categories.
\end{itemize}

\begin{figure}[!tb]
  \centering
  \includegraphics[scale=0.3]{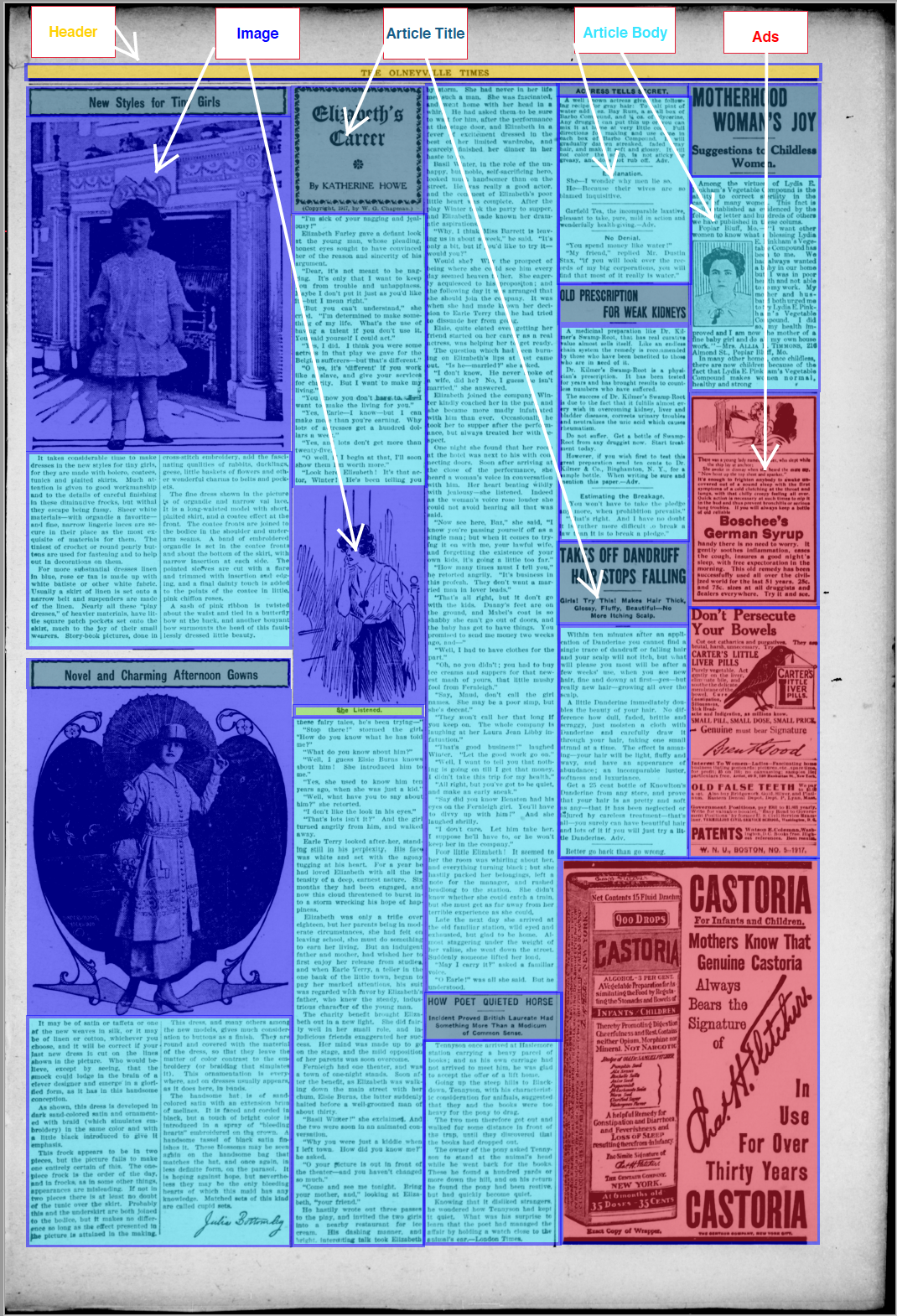}
  \caption{Example of Labeling Spec of NewsNet7 Dataset}
  \label{fig:NewspaperNet15}
 \end{figure}

\begin{table}[!tb]
\centering
\label{NewsNet7_data_stats}
\begin{tabular}{c|ccc}
\toprule
\multicolumn{1}{l}{}   & \multicolumn{3}{c}{\textbf{Number of instances}} \\ \hline
\textbf{Categories}    & \textbf{Train}  & \textbf{Test} & \textbf{Total} \\ \hline
\textbf{image}         & 10,017           & 1,793          & 11,810          \\
\textbf{article title} & 43,707           & 7,849          & 51,556          \\
\textbf{article body}  & 47,333           & 8,434          & 55,767          \\
\textbf{advertisement} & 35,165           & 6,228          & 41,393          \\
\textbf{table}         & 4,589            & 929            & 5,518           \\
\textbf{header}        & 1,930            & 343            & 2,273           \\
\textbf{other}         & 1,038            & 153            & 1,191           \\
\textbf{total}         & 143,779          & 25,729         & 169,508         \\
\bottomrule
\end{tabular}
\caption{NewsNet7 Dataset Statistics }
\end{table}

An illustration of page layout segmentation annotation is shown in Figure \ref{fig:NewspaperNet15}.
Table 1 summarizes the number of instance each category appears in the NewsNet7 dataset and how the dataset is split into train and test.

\paragraph{RDCL2019 (ICDAR2019 Competition on Recognition of Documents with Complex Layouts)}
The NewsNet7 dataset does not include a text ground truth which requires an extensive manual labor. In order to evaluate the impact of the layout segmentation algorithms on the downstream OCR results, the RDCL2019 layout analysis dataset's \cite{DBLP:conf/icdar/ClausnerAP19} example set is used, which includes 15 representative images with ground truth elements in XML format. It is important to note that these images are from contemporary documents such as magazines and they are only used to evaluate OCR performance. However, considering the fact that the layouts of these images are very different from historical newspapers, applying page segmentation models on this dataset as part of the process to obtain OCR results helps to assess their generalization capability.

\section{Newspaper Digitization Methods}
\begin{figure}[!b]
   \centering
\includegraphics[scale=0.45]{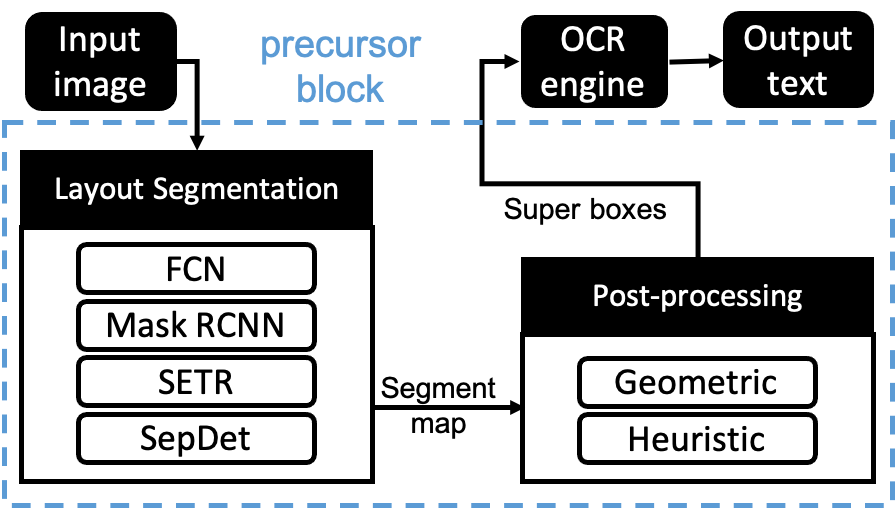}
   \caption{The framework pipeline}
   \label{fig:block_diagram}
 \end{figure}

The overall system architecture is illustrated in Figure \ref{fig:block_diagram} and can be summarized as follows. First, the input is an image of a document of arbitrary size, one of the listed document layout segmentation methods is applied to produce a segmentation map. Next, one of the listed post-processing methods is applied to convert the segmentation map to initial bounding boxes, then combined and ordered into super boxes. Then the super boxes are used to crop the image into high-resolution blocks passed to an OCR engine to extract the text.

\subsection{Document Layout Segmentation}
\begin{figure*}[!tb]
   \centering
   \includegraphics[width=\linewidth]{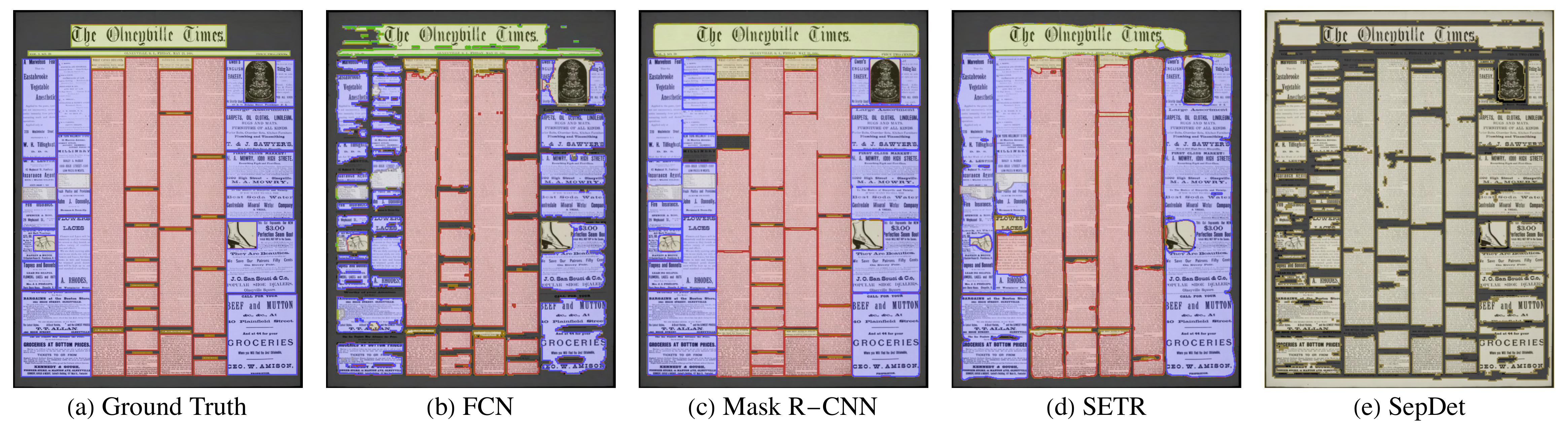}
   \caption{Comparison of example page layout segmentation results}
   \label{fig:all_vis}
 \end{figure*}

This section describes three different deep neural network architectures that are explored for document layout segmentation: Fully Convolutional Network (FCN), Mask R-CNN, and segmentation transformer (SETR). Figure \ref{fig:all_vis} shows the output of these models on a test newspaper image example.

\subsubsection{FCN}
Inspired by a related study \cite{DBLP:conf/icdar/MeierSSAC17} that performs semantic segmentation approach based on the visual appearance of the document, the FCN semantic segmentation model is re-implemented in this work as a baseline. The fully convolutional neural network includes feature extraction, down-sampling convolution, upscaling deconvolution, refinement convolution, and a classification layer that classifies the pixels. The model is trained from scratch with random weights initialization.

In the last layer of the model, two different class settings are considered for pixel classification: seven-class setting that considers each of the semantic content types (i.e. header, article title, article body, image, ad, table) as separate classes and a two-class setting that only regards foreground and background classes where foreground class includes all of the six content types; these settings are referred to as FCN-7 and FCN-2, respectively.  
The input images and the ground-truth masks are resized to 512x512 during the pre-processing steps
and they are randomly combined into batches of 20 samples to be fed into the neural network. A weighted cross entropy loss is applied to better address and minimize the effects of class imbalance on the model. The FCN model is trained for 100 epochs, which takes almost 6 hours on a single GPU.

\subsubsection{Mask R-CNN}
Mask R-CNN ~\cite{DBLP:conf/iccv/HeGDG17} is a widely-used state-of-art approach for instance segmentation. Since Mask R-CNN model predicts both bounding boxes and semantic masks, this approach saves us a significant extra effort to obtain bounding boxes for OCR engine. Furthermore, all the rest segmentation methods discussed in this paper require morphological processes and labeling components to convert the semantic mask into bounding boxes. Moreover, the semantic mask from Mask R-CNN could be either polygon, which would also work for layout contains more complex shape that rectangles could not represent.
The Detectron2 framework ~\cite{wu2019detectron2} provides an extensive collection of baseline models including Mask R-CNN that makes experimenting with this model a fast and friction-less process.

In this work, we used two different backbone structures in the architecture of Mask R-CNN: the first backbone is a Residual Networks (ResNet) combined with a feature pyramid network (FPN) structure that uses standard convolutional and fully connected heads for feature extraction. The second structure, Dilated-C5 (DC5), is a ResNet conv5 backbone with dilation in conv5 and convolutional and fully connected heads for mask and bounding box prediction.
Moreover, for the ResNet component, both ResNet-50 and ResNet-101 structures are used. Therefore, three different backbones are explored with Mask R-CNN, 
which are R50-FPN, R101-FPN, and R50-DC5. The backbones are initialized with Microsoft Research Asia's original models trained on COCO \cite{DBLP:conf/eccv/LinMBHPRDZ14} dataset. 
The input images are resized such that their shorter sides are 800 pixels. The models are trained with a mini-batch of 2 images per GPU. The model with R101-FPN backbone is trained on 4 GPUs for 800 epochs, and the ones with R50-FPN and R50-DC backbones are trained on 8 GPUs for 1000 and 600 epochs, respectively. With this training setup, each experiment takes around 3-4 days. The learning rate is set to 3e-4 for all of the experiments.

\subsubsection{SETR}
The Segmentation transformer (SETR) model (Zheng et al. 2021) has been recently shown to achieve the state of the art results in semantic segmentation on a number of datasets, including ADE20K and Pascal Context.

What makes SETR different from the previous segmentation models with fully convolution encoder-decoder structures is that it reformulates the task of semantic segmentation from a sequence to sequence point of view. Following this intuition, it considers the semantic segmentation task as translating an input image into a probability density map with the same resolution and employs an architecture similar to language translation models to tackle the problem.

To convert an input image into a sequence, we divide the image into a grid of fixed-size patches. Each patch is then flattened and linearly projected into an embedding space.  After adding positional encoding to the embedded vectors, they are fed to a transformer encoder that outputs a sequence of feature maps. Finally, a simple convolution decoder followed by some upsampling operations is applied to the feature maps to convert them to the original image resolution.

In our experiments with the SETR model, multiple different augmentation techniques such as random-cropping, random-resizing and flipping were used to make the training more robust to over-fitting. Moreover, 
a weighted cross-entropy loss is used for optimization to compensate for the class imbalance in input pixel distributions.
The model is trained with a batch size of 2 images for 1000 epochs. The input resolution is set to 512x512 pixels to allow training on a single GPU. The initial learning rate is set to 0.001 and it is decayed with a rate of 0.999 after every epoch. Under the above setting, training this model takes about 5 hours.

A series of heuristic post processing steps are applied to prepare the model's output for the next steps down the pipeline which will be explained in the next section.

\subsubsection{SepDet}
\begin{figure}[!b]%
\centering
\subfloat[\centering Predicted separators ]{{\includegraphics[width=.4\linewidth]{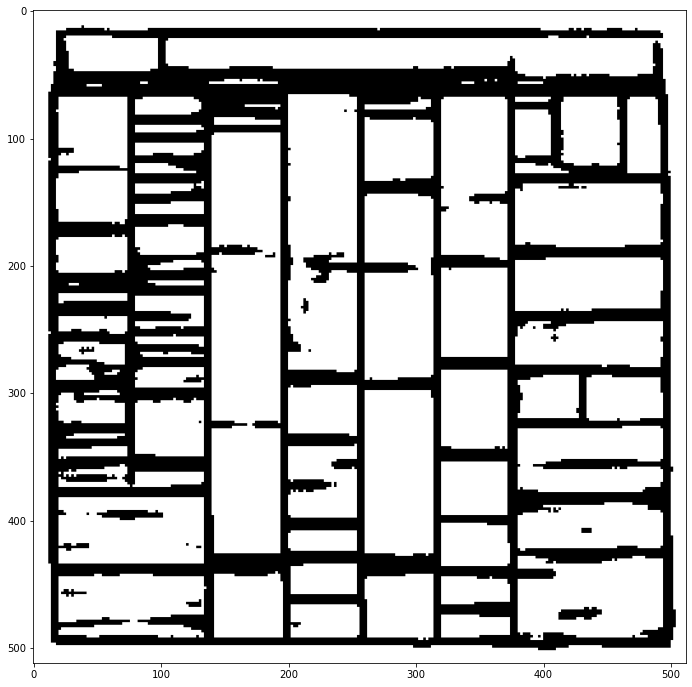} }}%
\qquad
\subfloat[\centering Ground truth of the separators ]{{\includegraphics[width=.4\linewidth]{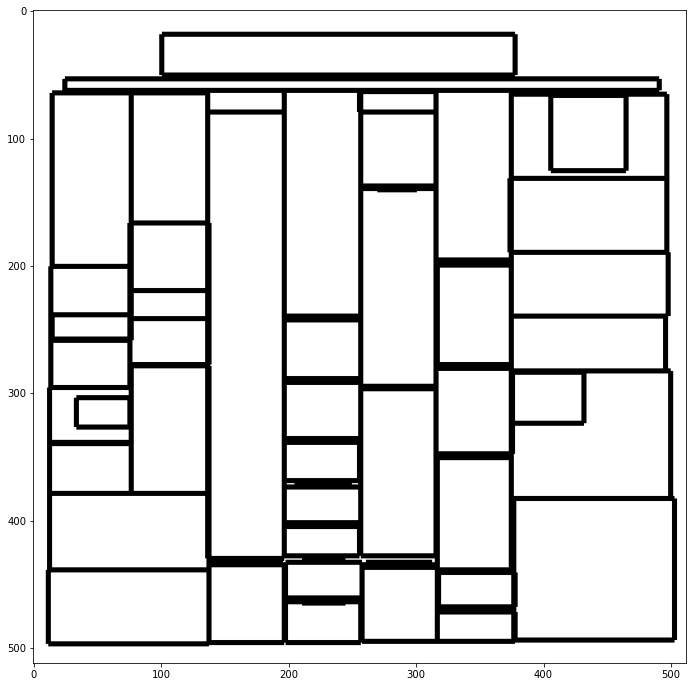} }}%
\caption{Predicted separators compared with ground truth}%
\label{fig:SepDetmodelPred}%
\end{figure}

Instead of putting the model focus on the entities that need to be digitized, a new deep neural net based model that followed the idea behind classic top-down approach  was proposed and experimented. The Separator Detector (SepDet) model was developed to detect separators (e.g. lines or white spaces that separate different articles or multiple columns of a single article) that exist in the documents. These separators are very useful for layout analysis as they contain important information to identify the logical structure of a newspaper page. The purpose of this model is to separate the text regions from the background and split them into a series of logical blocks using detected separators. Instead of using heuristics to identify the locations of these separators, a FCN model is trained to do such task.  As the model name indicates, this model does not classify the text regions to different entity types (e.g. table, ad and articles). For this model, the ground truth only contains 2 classes: separators and non-separators. To accelerate training while maintaining enough details for detecting the separators, the original input images are scaled to 512x512 pixels. As these separators often comprise less than 10 percent of the image area, a weighted cross-entropy loss is implemented for training the model. 
In our experiments with SepDet model, it is trained using a mini-batch size of 20 images for 100 epochs. The learning rate is set to 5e-4. And it takes 1 hour and 20 minutes to train the model on a single GPU.
A visualization of the predicted separators by SepDet for a test image and the corresponding processed ground truth map are shown in Figure \ref{fig:SepDetmodelPred}.

\subsection{Post-Processing Methods}

In this section, two post-processing strategies are introduced. These techniques convert the output of a segmentation model to a series of bounding boxes with a format that is proper for the OCR engine.

\subsubsection{Geometric}
\begin{figure}[!tb]
   \centering
   \includegraphics[width=\linewidth]{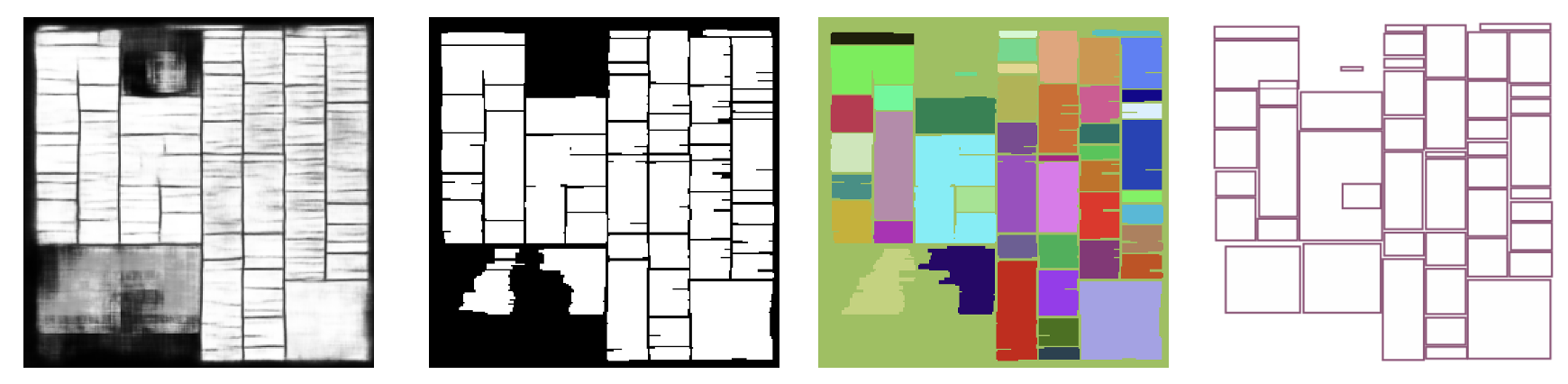}
   \caption{Illustration of the 4 steps to convert a semantic mask to bounding boxes, from left to right are original grayscale FCN-2 mask, binarized mask with morphological operations, connected components, and final bounding boxes}
   \label{fig:mask2bbox}
 \end{figure}

In this post-processing procedure, bounding boxes are computed and refined in the following steps.
First step is to extract bounding boxes from a predicted 2D array representing class probability distribution over pixels.
We started with applying Ostu’s thresholding method (N. Otsu 1979) to convert the probability map as a grayscale mask into a binary mask to tackle this task.
Then we applied the opening morphology operation to eliminate small regions in multiple iterations.
Next, we used the flooding algorithm to label all of the connected components distinctively. Finally, a bounding box is fitted to each refined connected component. Figure \ref{fig:mask2bbox} illustrates the explained process.

\begin{figure}[b]
   \centering
   \includegraphics[width=\linewidth]{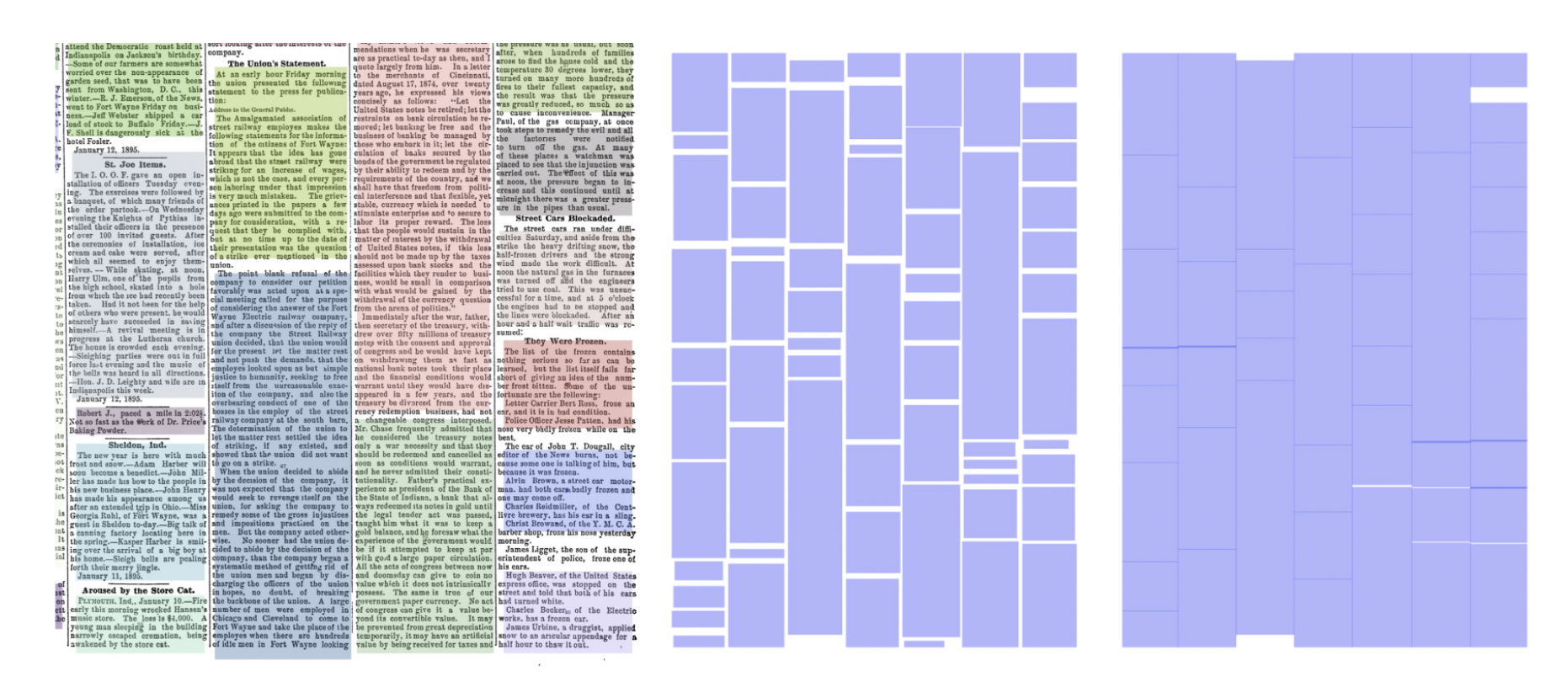}
   \caption{Left side: illustration of bounding boxes gap that will lead to character loss in OCR step; right side: illustration of clustering vertices operation.}
   \label{fig:character_loss}
 \end{figure}

The second step is removing gaps between bounding boxes.
There are often small gaps of several pixels between boundaries of annotated segments in the ground truth when the text columns are very close. Since several models are trained on down-sized ground-truth, and when scaling the bounding boxes to the original scale, those gaps can be inflated, resulting in missing text characters near the boundaries. 
To reduce those gaps, we expanded the predicted bounding boxes by applying a clustering algorithm (Figure \ref{fig:character_loss}) on bounding box vertices to replace the neighboring vertices with each cluster's centroid.

\subsubsection{Heuristic}

This post-processing algorithm serves three primary purposes. The first purpose is to convert the generated maps at the output of a segmentation model to bounding boxes. These bounding boxes are used to crop the input images around the text segments so that they can be digested by the OCR engine. The second purpose is to combine the generated bounding boxes by the previous step, whenever it is possible, to minimize the number of OCR API calls for cost optimization. Finally, the last purpose is to logically order the final bounding boxes so that the final extracted text can be read in the correct human-readable order.

In order to accomplish the first purpose, similar techniques explained in the geometric post-processing approach are employed.
To accomplish the second purpose, bounding boxes that are vertically aligned with some tolerance threshold are combined regardless of their class (as they are expected to be in the same column). Finally, to accomplish the third purpose, final combined bounding boxes are ordered from top left to bottom right to follow human reading habits.

\section{Experimental Results}
In this work, two different types of evaluation experiments have been conducted: isolated layout segmentation evaluation and end-to-end OCR evaluation. For all the models, the checkpoint corresponding to the best performance on the validation set was used for the experiments.
\subsection{Layout Segmentation Evaluation}

In order to assess the performance of the layout segmentation models, the predicted masks are evaluated using a series of pixel-wise metrics, including mean Intersection over Union (mIoU), mean Accuracy (mAc), mean Recall (mRe), mean Precision (mPr) and mean F-score (MFS). Each of these metrics was first evaluated for each class separately and then was averaged over all of the classes.

In the experiments, two different groups of models are considered, each with a different level of layout granularity. The first group of models included FCN-7, SETR, and Mask R-CNN models were trained to classify all of the seven classes on text regions. The second group of models, including the SepDet and FCN-2 models that were trained only to classify text and background pixels. The purpose of these experiments was to analyze how much having access to knowledge about different layout types can guide highly expressive models in the primary task of predicting text regions, which is the only required knowledge for the OCR engine. Due to this training methodology difference, these two groups are compared separately on their segmentation performance. Table  \ref{table:layoutmetric_7c}  
shows the segmentation performances of the first and second groups.

\begin{table}[h]
\small
\captionsetup{font=normalsize}
\begin{tabular}{l|cccccc}
\toprule
\textbf{model}  & \textbf{mIoU} & \textbf{mAc} & \textbf{mPr}  & \textbf{mRe}  & \textbf{mFs} & \textbf{IT}\\
\hline
101-FPN 	& 63.35 & 80.48 & 80.48 & 77.04 & 83.39 & 0.20\\ %

50-FPN  	& 63.90 & \textbf{80.66} & 80.66 & 77.45 & \textbf{83.82} & 0.15\\ %

50-DC5     	& 63.37 & 80.22 & 80.22 & 77.02 & 83.67 & \textbf{0.08}\\ %

SETR              	& \textbf{70.02} & 80.14 & \textbf{86.13} & \textbf{80.28} & 82.80 & 0.35 \\

FCN-7 &  56.34 & 71.38 & 74.08  & 70.99 & 79.60 & 0.49\\
\hline
\hline
FCN-2    &  \textbf{76.97}   & \textbf{84.92}  & \textbf{90.68}  & 86.78   & \textbf{88.57} & 0.49  \\
SepDet          & 44.01 & 83.23 &  46.27  &  \textbf{90.02}  &   61.12  & 0.30\\
\bottomrule
\end{tabular}
\caption{Layout segmentation performance for different segmentation models. The first 3 models are referring to Mask R-CNN. \textbf{IT} in this table represents inference time in seconds on a single NVIDIA V100 GPU. Note that SepDet and FCN-2 are evaluated on 2-class setting (foreground/background).}
\label{table:layoutmetric_7c}
\end{table}

When comparing the layout segmentation evaluation results from Table \ref{table:layoutmetric_7c}, we observed that the SETR and Mask R-CNN models perform very competitively in layout segmentation, and they considerably outperform the FCN model. It can also be observed that among different versions of the Mask R-CNN model, the one with R50-FPN backbone outperforms the rest and its performance is fairly close to the SETR model. Considering these observations while keeping the inference time of the models into account, the right choice of the model is highly dependent on the limitations of the use case. The SETR model and Mask R-CNN model with R50-FPN backbone are reasonable choices for those use-cases where the target application heavily relies on the accurate classification of layout segments. However, if the layout segmenting model is planned to be used as part of a real-time document processing pipeline, Mask R-CNN with R50-DC5 backbone is highly preferred since it has the lowest inference time while still delivering very promising segmentation performance. Finally if the application of interest is insensitive to correct classification of layout segments and it only relies on the final OCR results, FCN (2 classes) and SepDet models are preferred since they can separate the text blocks from the background with high accuracy while being fairly simple and easy to use.

\subsection{OCR Results}
In the previous ICDAR Page Segmentation competitions, PAGE format representation \cite{DBLP:conf/icdar/ClausnerAP17} was proposed for OCR ground truth. However, ground truth texts were often given in plain text instead of PAGE and XML format in practice. Herein, the OCR evaluation results are presented using methods that were based on ground truth texts in plain text format. These methods were built upon a one-to-one mapping between the predicted and ground-truth segments. First, for each word in the predicted segment, we locate its matching candidate's corresponding indices from the ground-truth; after that, given the fact that the words indices in a text segment must be continuous and coherent, we form a search space by enumerating all the possible candidates being continuous. Eventually, we select the candidate with the highest-ranking score by counting how many words are correctly matched. After such mapping was conducted, common metrics found in past studies were calculated, such as edit distance, read order and word recall.

\begin{itemize}
    \item \textbf{Edit distance:} The edit distance is a common similarity measure between two strings. It is defined as the minimum number of insertions, deletions or substitutions needed to transform one string to the other one.
    \item \textbf{Segment read order accuracy:} For newspaper digitization, maintaining proper read order is an essential factor. To evaluate this quality, a read order accuracy (ROA) metric is introduced as $ROA = 1 - m / n$, where $m$ is number of blocks out of order, and $n$ is total number of blocks; hence the higher the value (meaning more segments are in correct read order), the better. After each segment's matching interval is found in ground truth, the segments are sorted by their starting position to obtain the block order.
    \item \textbf{Word recall:} Word recall computes the percentage of words in ground truth that are correctly represented in the predicted text. The exact matches for uni-grams are counted, i.e.  each word is counted up to the number of times it appears in the ground truth. 
\end{itemize}

The above metrics are evaluated on RDCL2019 dataset and the results are shown in Table \ref{table:ocrmetric} which shows that SepDet-geometric approach results in the best read-order score while SETR-heuristic provides the best edit distance recall scores.
\begin{table}[h]
\centering
\footnotesize
\captionsetup{font=normalsize}
\begin{tabular}{c|ccc}
\toprule
\textbf{methods}  & \textbf{\begin{tabular}[c]{@{}c@{}}edit\\ distance\end{tabular}} & \textbf{\begin{tabular}[c]{@{}c@{}}read\\ order\end{tabular}} & \textbf{recall} \\
\hline

Baseline (no-layout-analysis) &  0.63  &  -   &  0.79 \\

SETR-heuristic &  \textbf{0.38}  &  0.82   &  \textbf{0.94} \\

FCN-7-heuristic   &  0.48  & 0.77  & 0.72 \\

FCN-7-geometric    & 0.47  & 0.76   & 0.74 \\

FCN-2-heuristic    & 0.51   & 0.79  & 0.66 \\

FCN-2-geometric    & 0.52  & 0.83  & 0.64 \\

SepDet-heuristic  &   0.515  &  0.65 & 0.88 \\

SepDet-geometric  & 0.62  & \textbf{0.98} & 0.79 \\

R101-FPN-heuristic & 0.42 & 0.68 & 0.86 \\

R101-FPN-geometric & 0.44 & 0.73 & 0.85 \\

R50-FPN-heuristic & 0.43 & 0.77 & 0.73 \\

R50-FPN-geometric & 0.46 & 0.81 & 0.72 \\

R50-DC5-heuristic & 0.52 & 0.70 & 0.86 \\

R50-DC5-geometric & 0.55 & 0.84 & 0.77 \\

\bottomrule
\end{tabular}
\caption{OCR metrics evaluated on the RDCL2019 example set. Note the last 6 models are referring to Mask R-CNN.}
\label{table:ocrmetric}
\end{table}

\section{Concluding Remarks}
In this paper, we proposed a precursor block to enhance existing OCR engines (e.g. Textract, Tesseract) for complex documents. Instead of feeding the entire scanned document into the OCR engine, the precursor block intelligently segments the document into smaller parts with similar semantic content and passes the segments to the OCR engine in an orderly fashion to preserve read-order. We made three contributions: a diverse dataset of newspapers with carefully designed layout annotations, an exploration of several state-of-the-art deep learning techniques to showcase promising results on document layout segmentation and a careful analysis on the effects of segmentation on read-order preserving OCR. Our experiments showed that each of the investigated layout segmentation architectures can deliver promising results under specific target constraints including high accuracy, low inference time and simplicity. Based on these results, existing OCR engines can integrate our proposed precursor block of layout segmentation into their framework to better handle documents with complex layouts.

\bibliography{aaai22}

\end{document}